%% file: crowdcounting.tex
\def\year{2021}\relax
\pdfoutput=1
\documentclass[letterpaper]{article} % DO NOT CHANGE THIS
\usepackage{multirow}
\usepackage{aaai21}  % DO NOT CHANGE THIS
\usepackage{times}  % DO NOT CHANGE THIS
\usepackage{helvet} % DO NOT CHANGE THIS
\usepackage{courier}  % DO NOT CHANGE THIS
\usepackage[hyphens]{url}  % DO NOT CHANGE THIS
\usepackage{graphicx} % DO NOT CHANGE THIS
\usepackage{natbib}  % DO NOT CHANGE THIS AND DO NOT ADD ANY OPTIONS TO IT
\usepackage{caption} % DO NOT CHANGE THIS AND DO NOT ADD ANY OPTIONS TO IT
\usepackage[switch]{lineno}  %

\title{STNet: Scale Tree Network with Multi-level Auxiliator for Crowd Counting}

\author{Mingjie Wang$^{1,2}$, Hao Cai$^{2}$, Xianfeng Han$^{3}$,  Jun Zhou$^{4}$, Minglun Gong$^{1\ast}$\\
	~\\
\small{$^{1}$University of Guelph, ON, Canada}\\
\small{$^{2}$Memorial University of Newfoundland, NL, Canada}\\
\small{$^{3}$Southwest University, Chongqing, China}\\
\small{$^{4}$Dalian Maritime University, Dalian, China}\\
}

\begin{document}

\maketitle

\begin{abstract}
Crowd counting remains a challenging task because the presence of drastic scale variation, density inconsistency, and complex background can seriously degrade the counting accuracy. To battle the ingrained issue of accuracy degradation, we propose a novel and powerful network called Scale Tree Network (STNet) for accurate crowd counting. STNet consists of two key components: a Scale-Tree Diversity Enhancer and a Semi-supervised Multi-level Auxiliator. Specifically, the Diversity Enhancer is designed to enrich scale diversity, which alleviates limitations of existing methods caused by insufficient level of scales. A novel tree structure is adopted to hierarchically parse coarse-to-fine crowd regions. Furthermore, a simple yet effective Multi-level Auxiliator is presented to aid in exploiting generalisable shared characteristics at multiple levels, allowing more accurate pixel-wise background cognition. The overall STNet is trained in an end-to-end manner, without the needs for manually tuning loss weights between the main and the auxiliary tasks. Extensive experiments on four challenging crowd datasets demonstrate the superiority of the proposed method.
\end{abstract}

\input{introduction}

\input{related}

\input{methods}

\input{experiments}

\bibliographystyle{aaai}
\bibliography{egbib}
\end{document}

%% file: introduction.tex
\section{Introduction}

Crowd counting recently has drawn lots of attention from researchers due to its great importance in a wide array of real-world applications including video surveillance, public crowd monitoring, and traffic control. The main objective of crowd counting is to infer the number of people in congested images~\cite{jiang2020attention}. Despite the exploration of pioneer works~\cite{zhang2016single,sam2017switching,cao2018scale,li2018csrnet,liu2019context,jiang2020attention}, crowd understanding is still a challenging issue for scenes exhibiting drastic scale variations, density inconsistency, or complex background.

\begin{figure}[t]
	\begin{center}
	\includegraphics[width=\linewidth,height=7cm]{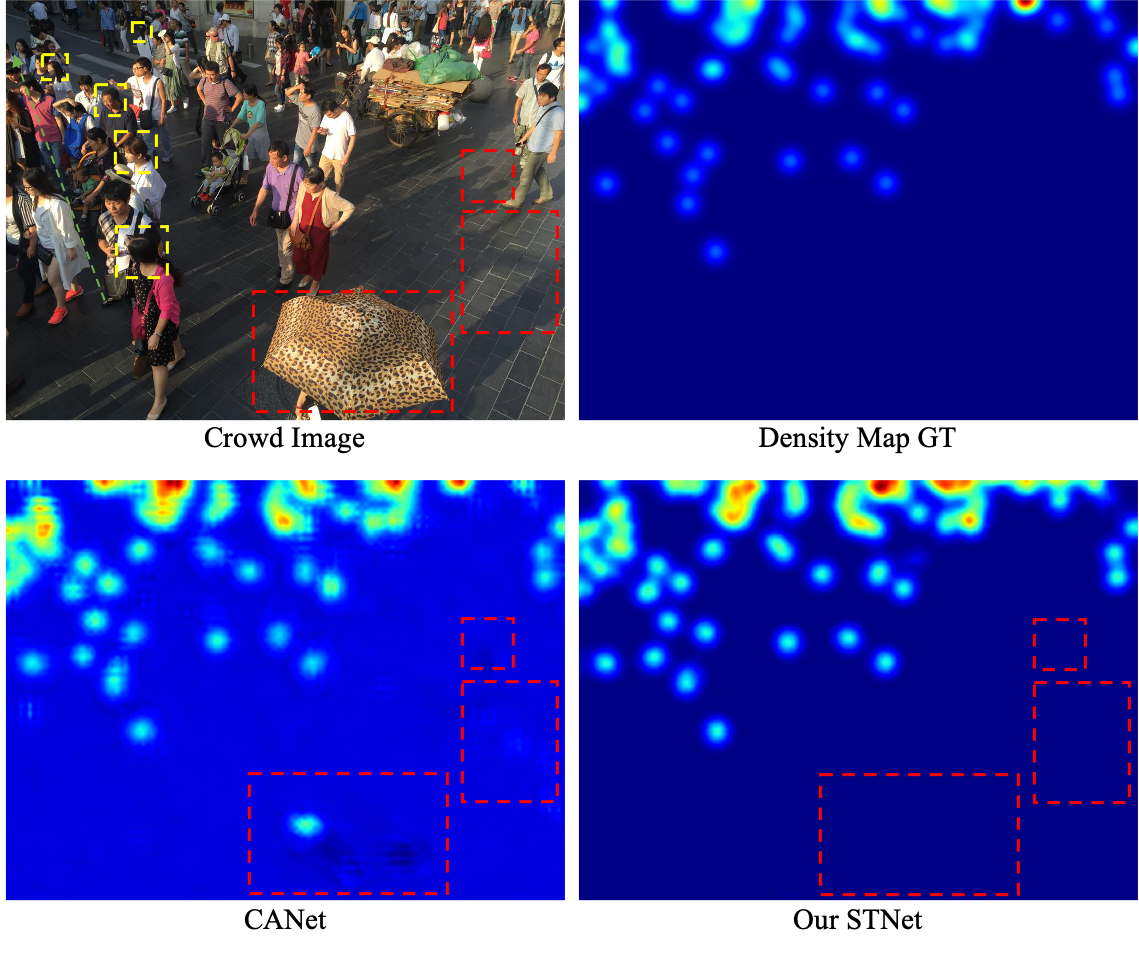}
	\end{center}
	\caption{Scale variations (yellow boxes) and complex background noises (red boxes) can often confuse existing approaches, such as CANet~\cite{liu2019context}. In comparison, our approach properly handles these areas.}
	\label{fig:problem_2}
\end{figure}

\begin{figure}[h]
	\begin{center}
		\includegraphics[width=\linewidth,height=5cm]{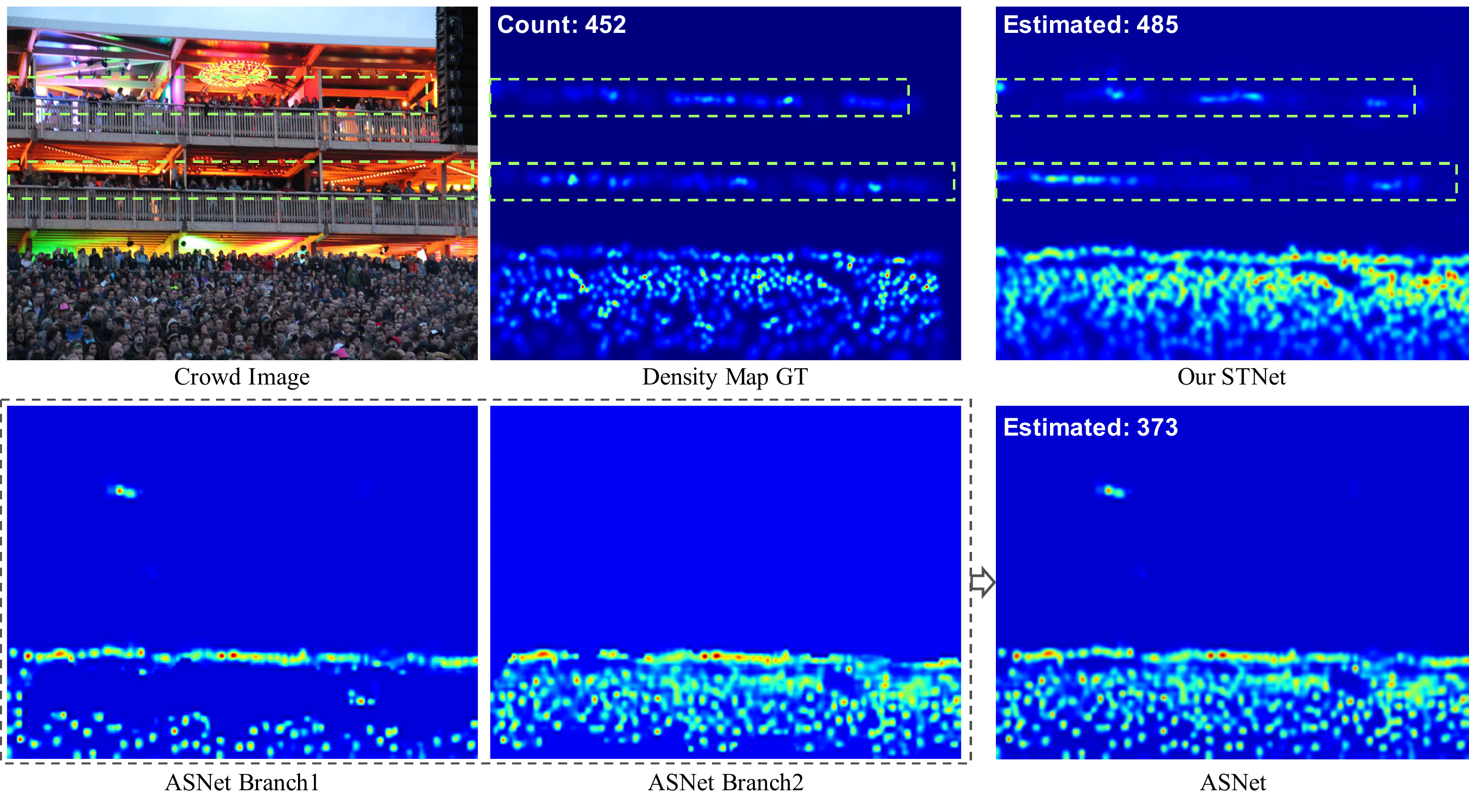}
	\end{center}
	\caption{Bounded by limited and discrete density levels, ASNet~\cite{jiang2020attention} fails to generate an accurate density map (bottom row) for the input scene from ShanghaiTech Part\_A~\cite{zhang2016single}. In comparison, the density map and counting result of our approach are much closer to the ground truth.} 
	\label{fig:problem_1}
\end{figure}

Along with the growing learning ability of deep neural networks, dozens of CNN-based solutions are sprung to tackle the aforementioned nuisances. The issue of scale variations (see Figure~\ref{fig:problem_2}) has attracted considerable attention from the research community~\cite{zhang2016single,sam2017switching,sindagi2017generating,cao2018scale,liu2019crowd,sindagi2019multi,liu2020adaptive,bai2020adaptive,jiang2020attention}. These methods endeavour to exploit multi-scale features through fusing representations from multiple layers~\cite{sindagi2019multi}, branches~\cite{bai2020adaptive} or columns~\cite{zhang2016single,sam2017switching} with varied receptive field sizes. 
However, all these methods only mine a limited range of scales in a coarse manner and hence do not perform well under wide scale diversity and uncertainty of scale variations.  Accordingly, an adequate scale diversity is desired to conform with the rapid intra- and inter-image scale changes of crowd heads or background objects.  

In fact, the procedure of scale parsing tends to be hierarchically structured (coarse-to-fine): smaller units (e.g., small heads)  are nested within larger units (e.g., foreground region)~\cite{shen2018ordered}, which are overlooked by most of existing methods.
S-DCNet~\cite{xiong2019open} attempts to utilize the spatial divide-and-conquer to decompose the counting. However, splitting the feature map into four subregions with the same area destroys the holistic geometric and semantic attributes of representations. The underlying structure for hierarchically parsing crowd has not been fully investigated in terms of fine-grained scale decomposition. Inspired by the promising application of tree structure in NLP~\cite{shen2018ordered}, we creatively integrate it to counting network for coarse-to-fine scale modeling, as well as for explicit expansion of scale range.

Density change across crowd scenes is also an open issue that causes accuracy degradation. While existing approaches~\cite{liu2019context,xiong2019open,sindagi2019multi,jiang2020attention} have tried to address this by re-weighing multiple branches with different density distributions, such approaches are prone to under- or over-fitting due to manually-designed levels and inconsistency of different branches. As shown in Figure~\ref{fig:problem_1}, even though two scale branches are utilized, ASNet~\cite{jiang2020attention} still cannot capture accurate people distributions in this example. The essence of these methods is to boost multiple columns to independently encode regions with different people densities. To alleviate the limitations of multi-branch-based methods, it is desired to impel a single-column network to gather hints that can accommodate density shifts.

%Supervision is known to be essentially crucial for the learning procedure of CNNs. Most current methods enforce the network to compute features at multiple scales in addition to density estimation by simultaneously minimizing the $L_2$ error of density map~\cite{liu2020adaptive}. The entangling supervision of scale-aware and density-aware features learning severely degrade the learning capability of the model due to completely inconsistent density and scale distributions. Specifically, the larger heads have smaller density values compared to smaller persons with larger density values~\cite{zhao2019leveraging}. Therefore, it is desired to disentangle the scale-focus supervision given the factor of dramatic scale variation to aid in featuring fine-grained scale information. Sindagi \emph{et al.}~\cite{sindagi2019multi} attempt to seek for auxiliary supervision to advance the combination of features from adjacent layers. However, its scale-aware groundtruth is created by splitting the density map into four categories and the whole network is optimized using Euclidean loss, which is still unable to cater for the need to disentangle scales from density values. 

As shown in Figure~\ref{fig:problem_2}, complex background is another factor that should not be neglected for accurate counting. Objects presenting in the background can vary dramatically (floor, umbrella, shadow, tree, etc.).  Hence, precise background filtering is vital to relieve pressure of density prediction.  
Considering the density-map-based paradigm, ground truth of density map is generated by convolving annotation map to indicate density values which are entangled with weak location information at each pixel. 
Therefore, the supervision of density map inevitably propagates the inherent ambiguity of signals, compounded by inconsistent numerical and semantic distributions, into learning procedure, which in turn negatively affects background filtering in the case of limited training set~\cite{gao2017deep}. Even though the SDANet~\cite{miao2020shallow} is empowered by focusing more on regions being crowd, the attention mechanism is optimized along with the main backbone through minimizing the $L_2$ error of density maps, which ignores the ambiguity problem. Another promising line of explorations~\cite{liu2019adcrowdnet,zhao2019leveraging,jiang2020attention} is to leverage auxiliary tasks (e.g., classification) to aid in the density estimation. It is observed that these methods bring minor benefits for extracting density- and background-aware characteristics. The deficiencies mainly arise from following reasons:

\begin{itemize}
\item Overlooking features from different levels impedes the encoding of fine-grained crowd regions. It is well-known that features from low-level layers encode details and spatial information, which can be exploited to achieve better localization~\cite{miao2020shallow}. Meanwhile, high-level representations encode global context~\cite{zhao2019leveraging,sindagi2019multi} with multi-scale information.  %SDANet~\cite{miao2020shallow} demonstrates the importance of shallow-layer features for localization details.

\item The insufficient relationship of knowledge from auxiliary stages (e.g. classification, segmentation, counting number and density estimation) causes the phenomenon that optimization directions of auxiliary branches are not conformable with each other~\cite{li2020dynamic}. 

\item There is a serious frequency imbalance of crowd and background pixels. For example, the percentage of background pixels is only 0.1867 in Part\_A and this observation is also supported by S-DCNet~\cite{xiong2019open}. The imbalance results in failing to capture discriminative features for the cognition of complex background. 

\item Finally, the training proceeds in a multi-stage manner and loss weights are manually tuned, which is tedious and prone to bias among side tasks. It is vital for multi-task learning to mine generalisable features. Loss supervisions should also enable learning of all tasks with equal contribution without allowing some tasks (e.g. density estimation) to dominate~\cite{liu2019end}.
\end{itemize}

\begin{figure*}[t]
	\begin{center}
		\includegraphics[width=0.9\linewidth,height=11cm]{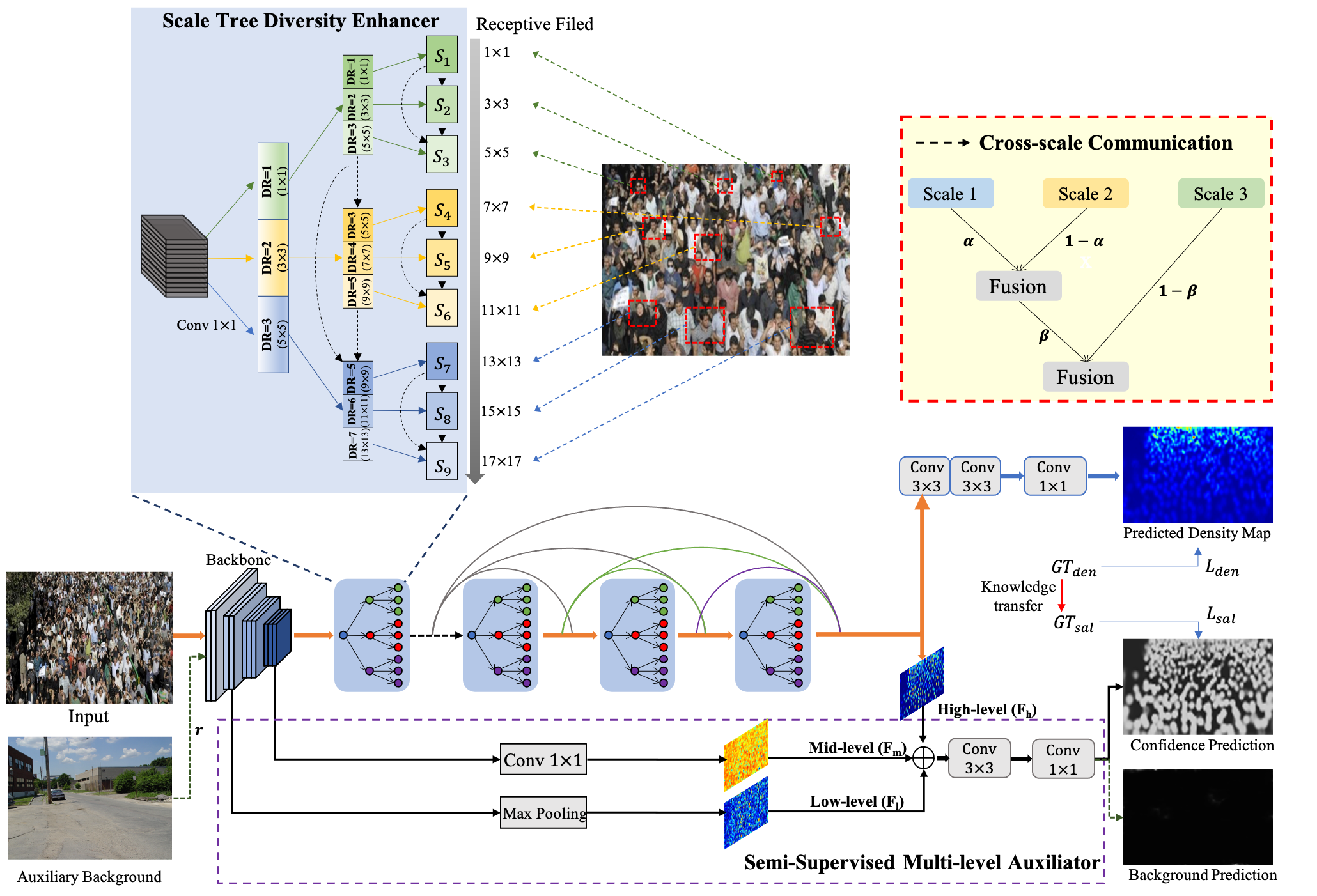}
	\end{center}
	\caption{Architecture overview of the proposed STNet. The network consists of two key components: Scale Tree Diversity Enhancer and Semi-supervised Multi-level Auxiliator. $\alpha$ and $\beta$ in cross-scale communication are random weights obeying the uniform distribution between 0 and 1. $r$ indicates the percentage of pure background images for semi-supervision and DR in the enhancer denotes the dilation rate. The whole network is trained in an end-to-end fashion.}  %\mlc{Why $S_4$ is $7\times7$, whereas DR=2 and DR=3 on its path are $3\times3$ and $5\times5$?} \mj{Yes.}
	\label{fig:network}
\end{figure*}

Towards the aforementioned issues, in this paper we coherently tackle wide-range scale variations, density inconsistency, and highly complex backgrounds for further performance improvement. To this end, we propose a novel \emph{Scale Tree Network} (STNet), see Figure~\ref{fig:network}, which is end-to-end trainable. Instead of simply stacking  parallel branches at multiple scales, we delicately design a parameter-efficient \emph{Scale Tree Diversity Enhancer} to widen the range of receptive fields and achieve the hierarchical parsing of coarse-to-fine scales. % (see Figure~\ref{fig:treemaps}).  

%\begin{figure}[t]
%	\begin{center}
%		\includegraphics[width=\linewidth,height=3.2cm]{treemaps.pdf}
%	\end{center}
%	\caption{The visualization of hierarchical parsing scales in a coarse-to-fine manner.} 
%	\label{fig:treemaps}
%\end{figure}

Furthermore, inspired by the success of multi-task learning~\cite{sindagi2017generating,zhao2019leveraging}, we reformulate the problems of density shift and complex background as a single \emph{multi-level confidence-driven attribute} instead of superfluous attributes which are unconformable with each other and heavily rely on manually specified density classification. This new form of attribute disentangles localization/global density distribution information inheriting from original density map. It also helps impose collaborative hints for crowd estimation on multi-level shared layers. Hence, we present a simple yet effective \emph{Semi-supervised Multi-level Auxiliator}, which aims at facilitating the  shared characteristics of the backbone CNN to embed crowd localization, holistic density distribution along with scale information.

Apart from novelty on network architecture, our  approach also differs in training procedure. Specifically, all of our proposed components are optimized jointly in an end-to-end manner. As shown in Figure~\ref{fig:problem_2}, the intrinsically cluttered background is usually mistakenly regarded as crowd due to the imbalance of background and crowd pixels, particularly in dataset with highly congested scenes. To address this issue, a semi-supervision strategy is incorporated when optimizing the auxiliary task through utilizing a number of {unlabelled and pure background images. Extensive experiments conducted on four popular crowd benchmarks (ShanghaiTech Part A and Part B~\cite{zhang2016single}, UCF\_QNRF~\cite{idrees2018composition}, UCF\_CC\_50~\cite{idrees2013multi}) demonstrate the superiority of our proposed method over state-of-the-art with best or competitive accuracy.

%% file: related.tex
\section{Related Work}
Recent years have witnessed the progress of crowd counting. The early approaches~\cite{zhao2003bayesian,sidla2006pedestrian,li2008estimating,subburaman2012counting} formulate the counting issue as a detection problem, which utilizes the hand-crafted low-level features to detect human parts. These approaches usually deliver poor results in congested images due to severe occlusion. Thus, density regression has been widely utilized to improve the counting accuracy, which maps features to density map~\cite{chan2008privacy,chen2013cumulative,chen2012feature,lempitsky2010learning,pham2015count}. Then the overall count is calculated by summing values over the predicted density map.  Thanks to the power of deep learning, CNN-based methods have become the mainstream of crowd analysis. We here review them based on their design objectives.

{\bf To handle scale variations.}  This group of approaches aim at mining multi-scale representations to address huge scale changes. MCNN~\cite{zhang2016single} leverages multiple columns to obtain features of multi-size receptive fields.  Hydra-CNN~\cite{onoro2016towards} learns a multi-scale non-linear regressor using a pyramid of input patches. Switch-CNN~\cite{sam2017switching} incorporates a side classifier to assign crowd patch to the appropriate estimation sub-model. Sindagi  \emph{et al.}~\cite{sindagi2017generating} propose to combine global and local contextual information for multi-scale features extraction. CSRNet~\cite{li2018csrnet} and ADCrowdNet~\cite{liu2019adcrowdnet} prove the effectiveness of dilated and deformable convolutions on expanding receptive fields. Recently, CAN~\cite{liu2019context} adaptively encodes the scale of the contextual information. MBTTBF~\cite{sindagi2019multi} employs a principled way of generating scale-aware supervision for training to increase the effectiveness of scale integration. DSSINet~\cite{liu2019crowd} employs CRFs to refine scale information with a message passing mechanism. Bai \emph{et al.}~\cite{bai2020adaptive} propose adaptive dilated convolution to adapt the scale variation. These approaches only consider limited scale diversity and discard the inherent hierarchical structure of parsing scale, which limits their ability to handle huge scale variations. 

{\bf To tackle changes of density distribution.} The common way to cope with density shifts is to learn rich representations covering adequate density levels. S-DCNet~\cite{xiong2019open} utilizes spatial divide-and-conquer strategy to recursively divide the dense scene into sub-regions with sparse crowd.
Jiang \emph{et al.} propose to use masks focusing on regions with different density levels to generate multi-branch density maps. DensityCNN~\cite{jiang2020density} jointly trains a density-level classifier to provide guidance. However, manually tuning the number of density levels is notorious.

{\bf To mitigate the effects of background noises.} Filtering noises (e.g. background pixels) in highly congested scenes has drawn growing attention from the community. One way to address this issue is to empower the network with attention mechanism for alleviating cluttered noises. Miao \emph{et al.}~\cite{miao2020shallow} try to diminish the impact of backgrounds via involving a shallow feature-based attention module. Another way is to employ auxiliary tasks. ADCrowdNet~\cite{liu2019adcrowdnet} views this issue as a classification task and trains an additional attention-aware network to detect crowd regions, whereas Zhao \emph{et al.}~\cite{zhao2019leveraging} formulate it as segment attribute attached on intermediate layers to guide the backbone to gain representations with latent semantic information. However, there is still much room for improvement in driving the backbone to capture more robust features in a multi-level fashion.

%% file: methods.tex
\section{Methodology}

%Our proposed framework of STNet is schematically shown in Figure~\ref{fig:network}. It consists of scale tree diversity enhancer and the semi-supervised multi-level enabler.  In this section, we elaborate on these two components.

\subsection{Scale Tree Diversity Enhancer}
With the drive to handle the challenges of polytropic head scales, a series of methods have been raised to acquire multi-scale features by resorting to multi-layer/branch/column aggregation with limited scale diversity. To further amplify the number of scales that a single output layer can represent and hierarchically parse coarse-to-fine scale information, we propose a delicately designed Scale Tree Diversity Enhancer. The Diversity Enhancer recursively divides the filter channels into small filter groups indicating diverse receptive filed sizes. These filter groups are organized in a tree structure, which allows reusing computation results and hence minimizes computational overhead. 

\begin{figure}[t]
	\begin{center}
		\includegraphics[width=\linewidth,height=3cm]{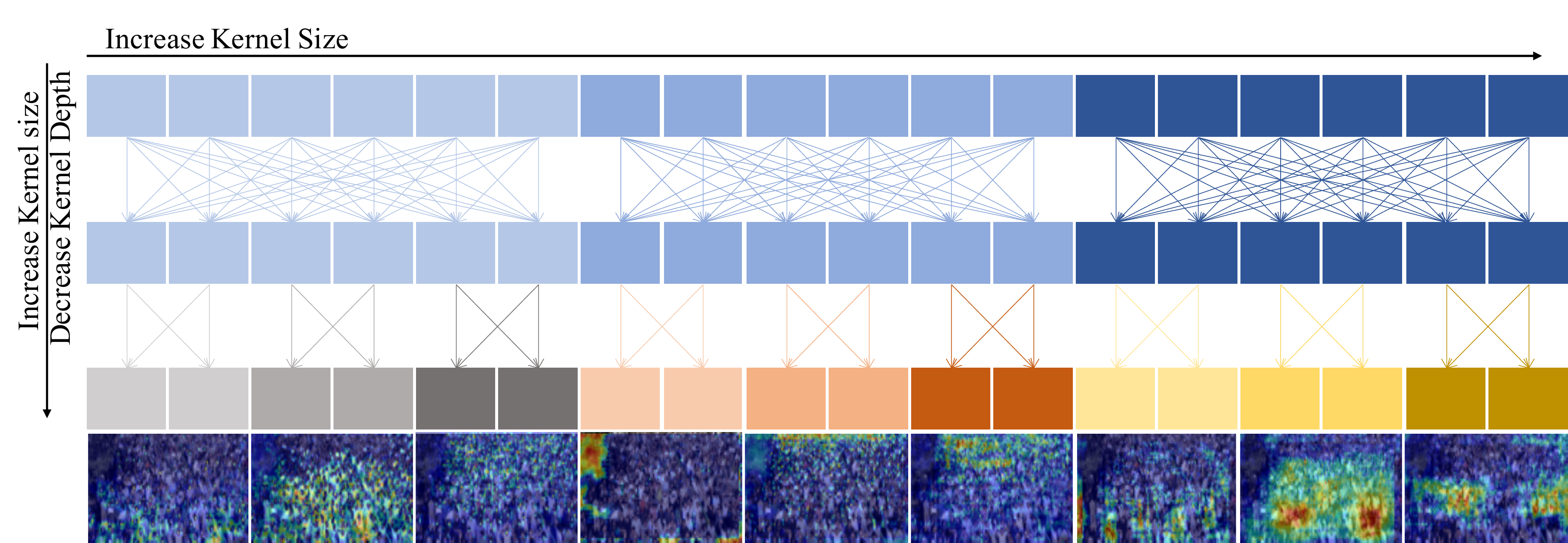}
	\end{center}
	\caption{The details of scale tree in our model. The channels are split recursively and the number of groups ranges from 1, 3 to 9. } 
	\label{fig:TSE}
\end{figure}

The pipeline of the proposed enhancer is shown in Figure~\ref{fig:network}. Following  previous work~\cite{li2018csrnet}, we adopt the first ten layers of pre-trained  VGG-16~\cite{simonyan2014very} as the backbone  to capture the low- and middle-level features. We then sequentially stack six scale tree diversity enhancers with dense connection for higher-level scale parsing.
As demonstrated in Figure~\ref{fig:TSE}, unlike most existing approaches that process scale information by a layer-wise way, our enhancer proceeds in multi-scale representations at a more granular level, which is based on group convolutions with kernel depths (number of input feature maps) and kernel sizes varying along two dimensions of channel and level.  The Diversity Enhancer first compresses the channel of input feature ($x$) via 1$\times$1 convolution $T_0$ and the output is viewed as the root of scale tree (parent scale $F_0$). Afterwards, the root features are split into three groups as nodes (children scales $F_1$) and then propagated to atrous convolution transformation $T_i$ with dilation rates $i$ in \{1, 2, 3\}. Recurrently, the nodes of second layer in the tree are deeply refined through dividing them into fine-grained leaves, followed by diverse sub-groups of dilated convolutions $T_j$ with dilation rates $j$ in \{1, 2, 3, 3, 4, 5, 5, 6, 7\}. To avoid the jumping change of receptive filed sizes, here dilation rates 3 and 5 are used twice. The terminal leaves (terminal scales $F_2$) involving abundant multi-scale information are concatenated into the next layer. Hence, the parsing procedure of the enhancer can be defined as:
\begin{equation}
\left\{
\begin{array}{lr}
F_0=T_0(W_0, X), &  \\
F_{1,n}=T_i(W_i,F_{0,n}),  n\in\{1,2,3\} &\\
F_{2,m}=T_j(W_j,F_{1,m}), m\in\{1,2...,9\},& 
\end{array}
\right.
\label{equ:tree}
\end{equation}
where $W_0$ indicates the trainable parameters of 1$\times$1 convolution $T_0$. $W_i$ denotes the path weights in the tree from parent scale $F_0$ to children scales $F_1$, whereas $W_j$ corresponds to the path weights from children scales $F_1$ to terminal scales $F_2$, respectively. The number of paths from root to leaves in our scale tree is used to characterize the scale diversity. In our implementation, the number of output scales is 9 and the maximum receptive filed size can reach to 17$\times$17.

{\bf Cross-scale Communication.}
Existing methods face the challenge of learning complementary scales on account of the ambiguity hiding in correlated adjacent scales~\cite{sindagi2019multi}. 
To avoid this issue and break the co-adaptation among children nodes in the scale tree, we design a mechanism of stochastic cross-scale communication. As shown in Figure~\ref{fig:network} (dotted box in red), the children scales $S_1$, $S_2$ and $S_3$ with receptive field sizes ranging from small to large ($S_1$ $<$ $S_2$ $<$ $S_3$) deliver message through recursive refinement. We formulate the communication as:
\begin{equation}
\hat{S_1}=S_1, \hat{S_2}=\alpha \hat{S_1}+(1-\alpha)S_2, \hat{S_3}=\beta \hat{S_2}+(1-\beta)S_3, 
\label{equ:cross_scale}
\end{equation}
where $\alpha$ and $\beta$ are random weights obeying the uniform distribution between 0 and 1. Before each training epoch, all variable gates are overwritten with new random values, while they are set to expected value of 0.5 during inference. By doing this, complementary scales are deeply exploited thanks to the reduction of ambiguity among children scales by stochastically blocking.  The additional advantage of cross-scale communication is to provide regularization cue for preventing overfitting.

{\bf Complexity Analysis.} Albeit tree-like structure seems to be complex, our scale enhancer is lightweight due to ingenious organization of various dilated convolutions.
The standard convolution usually contains a single type of kernel with kernel size $K^2$ and depth $D$. As for a standard computation block, it is comprised of one $1\times1$ conv layer and two layers of equal depths  kernel size $3^2$ and depth $D$. The number of parameters ($P$) and FLOPs can be computed as:
\begin{equation}
P=D^2+2\times9D^2=19D^2, FLOPs=19D^2WH,
\label{equ:stdconvcost}
\end{equation}
where $W$ and $H$ denote the spatial width and height of feature maps, respectively. Analogously, the number of parameters and FlOPs for our tree-based enhancer are:
\begin{equation}
P=D^2+3\times 9(\frac{D}{3})^2+9\times 9(\frac{D}{9})^2=5D^2
\label{equ:tsecost}
\end{equation}
\begin{equation}
FLOPs=5D^2WH
\label{equ:tseflops}
\end{equation}
With this formulation, our proposed module is more parameter-efficient ($5D^2$$<$$19D^2$) than multi-scale processing based on standard convolutions.

\subsection{Semi-supervised Multi-level Auxiliator}
%To further boost the learning of density- and background-aware characteristics, we reformulate the attributes of global density distribution and segment as a single multi-level confidence-driven attribute.

{\bf Multi-level Auxiliator.} To optimize for auxiliary tasks, previous approaches simply attach side supervisions to the top-most layer of feature extractor~\cite{zhao2019leveraging}. Due to optimization inconsistency between the main and auxiliary tasks, such approaches have limited benefits.

As we know, different depths represent distinct levels of abstraction in deep learning. Features with higher resolutions learned in shallow layers have fine spatial localization but lack semantics~\cite{li2020dynamic}. Hence, we hypothesize that lower-level features are also important for discriminating foreground pixels. Hence, a Multi-level Auxiliator is designed for counting network to assist in refining coarse-to-fine features, see Figure~\ref{fig:network} (bottom). Three levels of representations ($F_l$, $F_m$ and $F_h$) with different resolutions are drawn from shallow, middle and high intermediate layers. Afterwards, low- and middle-level features are processed by max pooling and 1$\times1$ convolution to be aligned with high-level feature maps. Then summation operation is used to integrate them. The fused features are fed into the sequence of  3$\times$3, 1$\times$1 convolutions, and the \emph{sigmoid} non-linear activation is utilized to generate the confidence map with background removed. The supervision signals from auxiliary task are backforward split into three routes to collaboratively optimize semantic-specific features.  

\begin{figure}[h]
	\begin{center}
		\includegraphics[width=\linewidth,height=2.0cm]{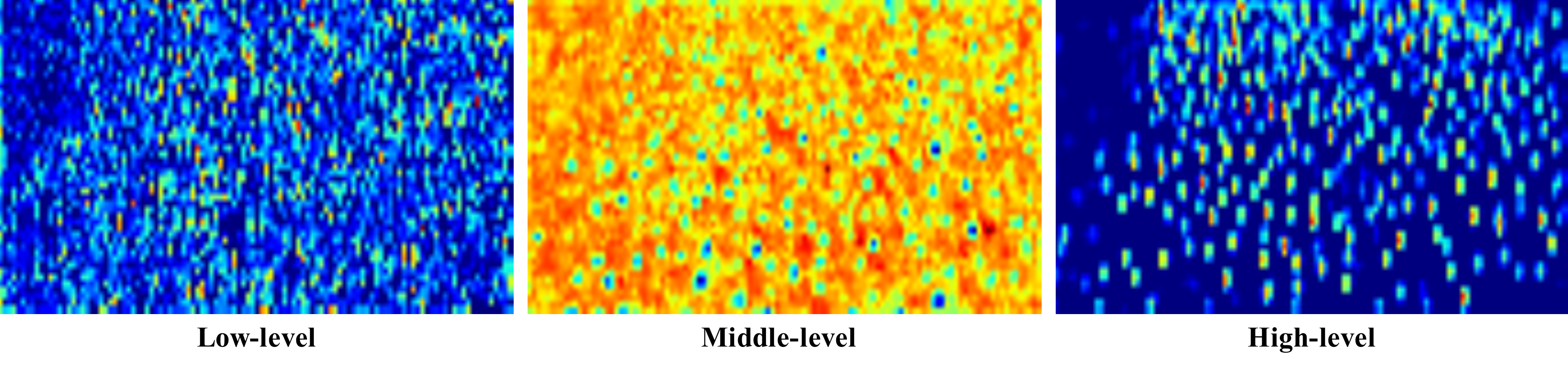}
	\end{center}
	\caption{Visualization of features at low, middle and high levels. The Multi-level Auxiliator guides the low-level features to involve more details compared to high level with abstract density distribution embedded, whereas middle level concentrates on noisy background regions.} 
	\label{fig:multilevel}
\end{figure}

We visually demonstrate our hypothesis in Figure~\ref{fig:multilevel}, which shows that low-level features indeed focus on detailed localization, whereas high level encodes abstract semantics. Interestingly, middle-level representation yield higher response to background pixels. Thereby, the Multi-level Auxiliator is essential to our goal of driving features located at different depths to embed diverse cues of density distribution and background information.
 
 {\bf Semi-Supervised Optimization.}
We attribute the poor handling of background noises to the latent imbalance of pixels occupied by background and people heads in crowd datasets. With a view to forcing the network to capture discriminative features for robustly perceiving complex noises, a semi-supervision strategy is incorporated to enlarge the background samples. As illustrated in Figure~\ref{fig:network}, a certain percentage rate ($\lambda$) of pure background images are stochastically inserted into the input batch. Through non-linear transformation of the backbone network, predicted density map, crowd confidence, and background maps are generated. Therefore, our holistic supervision $L$ consists of three sub-terms: $L_d$ and $L_c$ for supervised density estimation and confidence prediction, and $L_b$ for unsupervised background cognition. Then the overall optimization objective of our STNet is defined as:
\begin{equation}
L=L_d(W_d,I_c)+L_c(W_c,I_c)+L_b(W_c,I_b),
\label{equ:loss}
\end{equation}
where $I_c$ and $I_b$ indicate the inputs of congested scenes and background images, whereas $W_d$ and $W_c$ denote the sets of weight matrices collected from core density estimation and side confidence prediction, respectively. The first term is calculated using conventional Euclidean distance between the predicted density map ($D_d(W_d,I_c)$) and density ground truth ($D$). Let $N$ denote batch size, then
\begin{equation}
L_d(W_d,I_c)=\frac{1}{(1-\lambda)N}\sum_{i=1}^{(1-\lambda)N}||D_d(W_d,I_{c,i})- D_{i}||^2_2
\label{equ:euclidean}
\end{equation}

The second term $L_c(W_c,I_c)$ in loss function~(\ref{equ:loss}) is for the Auxiliator, which is implemented utilizing binary cross entropy between the predicted confidence map ($\hat{D}_d(W_d,I_c)$) and crowd label map $C$. We have:
%\small
%\begin{equation}
%L_c(W_c,I_c)=\frac{1}{(1-\lambda)N}\sum_{i=1}^{1-\lambda}\sum_{(p,q)\in I_(c,i)}(C_i^{p,q}logD_d(W_d,I_{c,i})^{p,q}+(1-C_i^{p,q})log(1-D_d(W_d,I_{c,i})^{p,q}))
%\label{equ:crossentropy}
%\end{equation}
{\small
\begin{equation}
\begin{array}{lr}
L_c(W_c,I_c)=\frac{1}{(1-\lambda)N}\sum_{i=1}^{(1-\lambda)N}\sum_{(p,q)\in I_{(c,i)}} &  \\
~&\\
(C_i^{p,q}log\hat{D}_d(W_d,I_{c,i})^{p,q}+(1-C_i^{p,q})log(1-\hat{D}_d(W_d,I_{c,i})^{p,q}))&
\end{array}
\label{equ:crossentropy}
\end{equation}
}

In order to combat the inconsistency of two optimization directions and consider global density level at the same time, we derive crowd label by binarizing from density map $D$ in terms of the average density values.
\begin{equation}
D_{p,q}=\left\{
\begin{array}{lr}
1 & if~\hat{D}_{p,q} >= 0.5\times Mean(D)\\
~&\\
0 &  otherwise,
\end{array}
\right.
\label{equ:crowdlable}
\end{equation}
where the purpose of $Mean(D)$ is to embed holistic density information of crowd images into auxiliary deep supervisions. The last term $L_b(W_c,I_b)$ is designed for noise-focusing optimization without additional labels on the basis of $L_2$ norm of predicted background $B(W_c,I_b)$.
\begin{equation}
L_b(W_c,I_b)=\frac{1}{\lambda N}\sum_{i=1}^{\lambda N}||B(W_c,I_{b,i})||^2_2
\label{equ:semiloss}
\end{equation}

To guarantee learning of two tasks with equal contribution and get rid of cumbersome weights tuning, all optimization terms are treated equally with the same weight of ``1''. %thanks to the consistency of delicately designed enabler.

%% file: experiments.tex
\section{Experiments}

\subsection{Experiment Setup}
%{\bf Implementation Details.} 
The first ten layers of a pretrained VGG-16 are leveraged to form the backbone in STNet, followed by six diversity enhancer modules with dense connections. Each scale tree has depth $=2$ and degree $=3$. Low-and middle-level supervisions are imposed on the fifth and top layers in backbone, respectively.  
During each epoch, we augment the input crowd samples by cropping fixed-size patches of {\small176$\times$176} at random locations and then the image patches are randomly flipped horizontally and vertically. In addition, random color jitter is applied to enlarge the training set. Different from other models seeking for auxiliary tasks, the training pattern of our method is completely end-to-end. The Adam algorithm with initial learning rate of 0.0001 is used to optimize our model and the batch size is uniformly set to 16. The pure background samples for semi-supervision training procedure are 675 images without people involved, which are taken from Internet and are used for all 4 training datasets. After the training, the Multi-level Auxiliator is removed and the main stream of density estimation with strong representing ability is used to infer density maps. The proposed STNet is implemented in Pytorch~\cite{paszke2017automatic} and  all experiments are carried out on a single Titan Xp GPU.

%{\bf Datasets.} {\bf ShanghaiTech} dataset~\cite{zhang2016single} consists of two parts: Part\_A and Part\_B. This dataset includes 1198 crowd images with a total of 330,165 annotated people. Part\_A contains 482 congested internet images while Part\_B has 716 sparse images with a fixed size of {768$\times$1024} captured from outside streets. {\bf UCF\_QNRF}~\cite{idrees2018composition} dataset is collected from the website and includes 1,553 images with a total number of 1,252,642 people, in which 1201 images are used for training and 334 are selected as testing set. This dataset has more diverse densities, backgrounds and perspectives. {\bf UCF\_CC\_50}~\cite{idrees2013multi} dataset contains 50 images taken from the Internet with an average of 1280 individuals per image. This is a very challenging benchmark due to the limited crowd scenes and huge variations in terms of crowd density. 
 %We utilize 5-fold cross validation to evaluate our method. 

%{\bf Evaluation Metrics.} Following the previous work~\cite{liu2020adaptive}, the mean absolute error (MAE) and root mean square error (MSE) are used as evaluation metrics, which are defined as follows:
%{\small
%\begin{equation}
%MAE=\frac{1}{N}\sum_{i=1}^{N}|C_i-C_{i}^{GT}|, MSE=\sqrt{\frac{1}{N}\sum_{i=1}^{N}||C_i-C_{i}^{GT}||^2}
%\label{equ:evaluation}
%\end{equation}
%}
%where $N$ is the number of testing crowd images and $C_i$ denotes predicted crowd count whereas $C_i^{GT}$ indicates the actual number of people presenting in the $i_{th}$ crowd image. 

\begin{table*}[h]
	\begin{center}
		\begin{tabular}{c|cc|cc|cc|cc}
			\hline
			\multirow{2}{*}{Methods}& \multicolumn{2}{c|}{{\bf Part\_A}} 
			&\multicolumn{2}{c|}{{\bf Part\_B}}&\multicolumn{2}{c|}{{\bf UCF-QNRF}}&\multicolumn{2}{c}{{\bf UCF\_CC\_50}}\\
			\cline{2-9}
			~& MAE & MSE & MAE & MSE  & MAE & MSE  & MAE & MSE\\
			\hline
			TEDNet{\small \emph{(CVPR19)}} & 64.2 & 109.1 & 8.2 & 12.8  & 113 & 188 & 249.4 & 354.5\\ %~\cite{jiang2019crowd} 
			ADCrowdNet {\small \emph{(CVPR19)}} & 63.2 & 98.9 & 7.6 & 13.9  & - & - & 257.1 & 363.5\\ %~\cite{liu2019adcrowdnet}
			PACNN + CSRNet {\small \emph{(CVPR19)}}& 62.4 & 102.0 & 7.6 & 11.8  & - & - & 241.7 & 320.7\\ %~\cite{shi2019revisiting}
			CANet {\small \emph{(CVPR19)}} & 62.3 & 100.0 & 7.8 & 12.2 & 107 & 183 & 212.2 & 243.7\\ %~\cite{liu2019context}
			SPN+L2SM {\small \emph{(ICCV19)}} & 64.2 & 98.4 & 7.2 & 11.1  & 104.7 & 173.6 & 188.4 & 315.3\\ %~\cite{xu2019learn}
			MBTTBF-SCFB {\small \emph{(ICCV19)}} & 60.2 & 94.1 & 8.0 & 15.5 & 97.5 & 165.2 & 233.1 & 300.9\\ %~\cite{sindagi2019multi}
			DSSINet {\small \emph{(ICCV19)}} & 60.63 &	96.04  & 6.85 &	10.34 & 99.1	&	159.2 & 216.9 &	302.4\\ %~\cite{liu2019crowd}
			S-DCNet {\small \emph{(ICCV19)}} & 58.3 &	95.0  & 6.7	& 10.7 & 104.4	&	176.1 & 204.2	& 301.3\\ %~\cite{xiong2019open}
			ASNet {\small \emph{(CVPR20)}}& 57.78 & 90.13  & -	& - & 91.59	& 159.71 & 174.84&  251.63\\ %~\cite{jiang2020attention}
			ADNet {\small \emph{(CVPR20)}} & 61.3 & 103.9  & 7.6 & 12.1 & 90.1 & 147.1 & 245.4 & 327.3\\ %~\cite{bai2020adaptive} 
			AMRNet {\small \emph{(ECCV20)}} & 61.59 & 98.36  & 7.02 & 11.00 & 86.6 & 152.2 & 184.0 & 265.8\\ %~\cite{liu2020adaptive}
			AMSNet {\small \emph{(ECCV20)}} & 58.0 & 96.2  & 7.1 & 10.4 & 103 & 165 & 208.6 & 296.3\\
			\hline
			\hline
			BL {\small \emph{(ICCV19)}} & 62.8 & 101.8  & 7.7 & 12.7 & 88.7 & 154.8 & 229.3 & 308.2\\ %~\cite{ma2019bayesian}
			ADSCNet {\small \emph{(CVPR20)}} & 55.4 & 97.7  & 6.4 & 11.3 & {\bf 71.3} & {\bf 132.5} & 198.4 & 267.3\\ %~\cite{bai2020adaptive}
			\hline
			\hline
			{\bf STNet} (proposed) & {\bf 52.85} & {\bf 83.64} & {\bf 6.25} & {\bf 10.30}  & 87.88 & 166.44 & {\bf 161.96} & {\bf 230.39} \\
			\hline
		\end{tabular}
	\end{center}
	\caption{
		Comparison with state-of-the-art methods on ShanghaiTech (Part\_A and Part\_B), UCF-QNRF and UCF\_CC\_50 datasets. Best results are shown in boldface.
	}
	\label{table:comparision}
\end{table*}

%{\bf Dataset Used.} 
%\mlc{briefly explain where you get the datesets and why you picked them.  cite other papers using the same datasets.}

\subsection{Comparisons with State-of-the-art}
We first qualitatively compare the performance of the proposed STNet with recent prominent  approaches~\cite{jiang2019crowd,liu2019adcrowdnet,ma2019bayesian,bai2020adaptive,liu2020adaptive} on four datasets.  It is worth noting that some existing approaches profit from novel frameworks with conventional density map or new ground truths. Take BL~\cite{ma2019bayesian} and ADSCNet~\cite{bai2020adaptive} for example. They propose new types of ground truth to boost prediction accuracy. In this paper, we focus on investigating novel counting framework embracing higher representation capacity and all our results are acquired by training the model with original ground truth. It is more reasonable to compare with these frameworks where the loss is computed on the given ground truth of original density map. Qualitative results are shown in Table~\ref{table:comparision}. 

On both datasets of ShanghaiTech Part\_A and Part\_B, our model surpasses all existing methods and achieves the state-of-the-art result with MAE 52.85 and 6.25, which are lower than that of the latest ADSCNet. It should be noted that the success of ADSCNet (55.4 \& 6.4) (CVPR20) is mainly attributed to ground truth correction operation. When compared with its counterpart ADNet (61.3 \& 7.6) trained using plain density map, our model obtains greater improvements: 13.8\% and 17.8\% lower for Part\_A and Part\_B accordingly, which illustrates that our method has more powerful ability of representation learning in the case of either dense or sparse crowd datasets.
As for UCF-QNRF dataset, our method achieves the third-best result compared with all existing methods and the second-best result of MAE 87.88 among the set of approaches that use the original ground truth, which is only 1.5\% higher than  that of AMRNet (ECCV20). Note that AMRNet only gains slightly better performance on this dataset and reports relatively weak results on ShanghaiTech Part\_A and Part\_B. Benefiting from ground truth correction, ADSCNet obtains the best result on UCF-QNRF. For fair comparison, here we emphasize the comparison with its primary version ADNet. 
It can be observed that our STNet performs better (MAE 87.88) than ADNet (MAE 90.1), which demonstrates that our algorithm is able to excel in capturing informative features. Finally, on the tiny UCF\_CC\_50, our method also achieves the best result with an MAE of 161.96, which is 7.4\% lower than that of the second-best approach ASNet (CVPR20).

\begin{table}[h]
	\begin{center}
		\begin{tabular}{|c|c|c|c|c|c|c}  %{\hsize}{@{}@{\extracolsep{\fill}}ccccc@{}}
			\hline
			& CSRNet & CSRNet+(w/o Enhancer) & STNet \\
			\hline
			\hline
			Params.& 16.26M & 25.10M & 15.56M\\
			\hline
			MAE& 68.2  & 58.7 & 52.8\\
			MSE& 115.0 & 97.2 & 83.6 \\
			\hline
		\end{tabular}
		\caption{The impacts of Scale Tree Diversity Enhancer on ShanghaiTech Part\_A dataset.}
		\label{table:enhancer}
	\end{center}
\end{table}

\subsection{Ablation Study}
In this section, we conduct detailed ablation study on ShanghaiTech Part\_A to demonstrate the effectiveness of components in the proposed  STNet.

\noindent
{\bf Effect of Scale Tree Diversity Enhancer.} 
To investigate the effects of the Diversity Enhancer, we compare two networks: traditional CSRNet~\cite{li2018csrnet} and CSRNet+. CSRNet+ means the STNet without tree enhancer, which is implemented by replacing the enhancers in STNet with traditional CSRNet units. The quantitative results in Table~\ref{table:enhancer} show that the proposed Diversity Enhancer is remarkably better than the widely-used layer-wise dilated convolution in CSRNet, while introducing fewer parameters at the same time. The improvement is mainly attributed to the fact that the scale tree provides huger diversity of scales.

\noindent
{\bf Effect of Multi-level Auxiliator}. To evaluate the merit of Deep Auxiliator, we compare several variants of CSRNet and STNet. As shown in Table~\ref{table:enabler}, CSRNet+ with  Multi-level Auxiliator} achieves 5.9\% higher MAE than that without the designed auxiliator. In order to further investigate the impacts of features extracted from different levels, we evaluate four variants of STNet: from STNet without auxiliator to STNet with auxiliator using high-middle-low-level. The quantitative MAE values demonstrate that all three levels of the auxiliator contribute to the final performance. Especially, missing low-level information brings more dramatic decreases in MAE. This phenomenon further supports the observation that low-level features encode abundant detailed localization information, which is in favor of purifying crowd features. To combat the inconsistency of two optimization directions (density estimation \& crowd confidence) and take the holistic density distribution into account, we disentangle ground truth of crowd label from density map according to Equ.~\ref{equ:crowdlable}. To illustrate the effectiveness of this generation strategy, we compare it with the approach in the recent work~\cite{zhao2019leveraging} with binaryzation threshold value of 0 and our algorithm brings 14.19\% lower in MAE (61.59\% $\rightarrow$ 52.85\%).
\begin{table}[h]
	\begin{center}
		\begin{tabular}{|c|c|c|}  %{\hsize}{@{}@{\extracolsep{\fill}}ccccc@{}}
			\hline
			& MAE & MSE\\
			\hline
			CSRNet+ w/o Auxiliator & 62.378 & 100.951
\\
			CSRNet+ w/ Auxiliator &  58.711 & 97.197\\
			\hline
			STNet w/ BT=0 & 61.591 & 100.313\\
			STNet w/o Auxiliator &  59.296 & 98.699\\
			STNet w/ H.A & 57.66 & 98.93\\
			STNet w/ H.M.A & 57.519 &98.86\\
			STNet w/ H.M.L.A &  52.85 & 83.64\\
			\hline
		\end{tabular}
		\caption{Multi-level Auxiliator ablation study on ShanghaiTech Part\_A. H.A, H.M.A and H.M.L.A indicates the auxiliator uses high-level only, high and middle-level only, and all three levels, respectively.}
		\label{table:enabler}
	\end{center}
\end{table}

\noindent
{\bf Effect of Semi Supervision with Pure Background Samples.}  Table~\ref{table:semi} shows quantitative comparisons between the STNet with and without using pure background images for training the Multi-level Auxiliator. It is observed that additional background samples effectively boost the capacity of capturing representations encoding background information, thereby leading to precise recognization of pixels occupied by people.
   
\begin{table}[h]
	\begin{center}
		\begin{tabular}{|c|cc|cc|}  %{\hsize}{@{}@{\extracolsep{\fill}}ccccc@{}}
			\hline
			&\multicolumn{2}{|c|}{{w/o Semi Supervision}} &\multicolumn{2}{|c|}{{w/ Semi Supervision}}\\
			\cline{2-5}
			& MAE & MSE & MAE & MSE \\
			\hline
			Part\_A& 56.47  & 97.19 & 52.85 & 83.64\\
			Part\_B& 6.71 & 11.36 & 6.25 & 10.30\\
			QNRF & 89.81 & 168.29 & 87.88 &166.44\\
			\hline
		\end{tabular}
		\caption{Semi Supervision ablation of our STNet on ShanghaiTech Part\_A, Part\_B and UCF\_QNRF datasets.}
		\label{table:semi}
	\end{center}
\end{table}

\section{Conclusion}
In this paper, we present a novel network, STNet, which consistently addresses the challenges of drastic scale variations, density changes, and complex background.  The issues of scale variations are alleviated using a novel tree-based scale enhancer, whereas a Multi-level Auxiliator is designed to accurately filter out pixels from complex background and adapt to density changes. Extensive experiments on four popular crowd counting datasets demonstrate the superiority of our STNet over the state-of-the-art approaches, while employing fewer parameters at the same time. A simple yet effective idea of using pure background images to deal with crowd and background sample imbalance problem is also introduced. This idea can be easily incorporated in other crowd counting algorithms for further accuracy improvement. For future work, we will investigate possibility of using self-supervision and more unlabelled scenes to improve crowd counting performance.